\documentclass[letterpaper, 10 pt, journal, twoside, final]{ieeetran}

\IEEEoverridecommandlockouts                              

\expandafter\let\csname equation*\endcsname\relax
\expandafter\let\csname endequation*\endcsname\relax

\title{Geometrically modulable gait design for quadrupeds}

\author{Hari Krishna Hari Prasad$^{1}$ , Ross L. Hatton$^{2}$, and Kaushik Jayaram$^{1,*}$
\thanks{Manuscript received: February, 29, 2024; Revised April, 25, 2024; Accepted June, 11, 2024.} 
\thanks{This paper was recommended for publication by Editor Abderrahmane Kheddar upon evaluation of the Associate Editor and Reviewer's comments.}
\thanks{$^{1}$Animal Inspired Movement and Robotics Laboratory, Paul M. Rady Department of Mechanical Engineering, University of Colorado Boulder} 
\thanks{$^{2}$ Mechanical, Industrial and Manufacturing Engineering, Oregon State University}
\thanks{$^{*}${For correspondence, \tt\footnotesize kaushik.jayaram@colorado.edu}}%
\thanks{Code: \href{https://github.com/Animal-Inspired-Motion-And-Robotics-Lab/Paper-Geometrically-modulable-gait-design-for-Quadrupedal-robots}{\emph{Codebase-geometrically-modulable-gaits}}}
\thanks{Digital Object Identifier (DOI): see top of this page.}
} 

\IEEEoverridecommandlockouts
\usepackage{mathnotation}
\usepackage{amsmath,amssymb,amsfonts}
\usepackage{esint}
\usepackage{tensor}
\usepackage{algorithmic}
\usepackage{graphicx}
\usepackage{graphics}
\graphicspath{{./Figures/Iter2/}}
\usepackage{textcomp}
\usepackage{xcolor}
\usepackage{gensymb}
\usepackage[utf8]{inputenc}
\usepackage{times}
\usepackage{mathtools}
\usepackage{cancel}
\usepackage{subcaption}
\usepackage[separate-uncertainty=true,list-units=repeat,multi-part-units=single]{siunitx}
\usepackage{nicefrac}
\usepackage{multicol}
\usepackage{multirow}
\usepackage{array}
\usepackage{pgfplots} 
\usepackage{kantlipsum}
\usepackage{setstack}
\pgfplotsset{compat=1.14}
\usepackage{float}
\usepackage{stfloats}
\usepackage[flushleft]{threeparttable}
\usepackage{comment}
\usepackage{enumitem}
\usepackage{todonotes}

\usepackage{titlesec}
\titlespacing*{\section}{5pt}{0.1\baselineskip}{0.2\baselineskip}
\titlespacing*{\subsection}{1pt}{0.1\baselineskip}{0.2\baselineskip}

\newlength{\imagewidth}

\usepackage[numbers]{natbib}
\usepackage[pageanchor=true,plainpages=false, pdfpagelabels, bookmarks,bookmarksnumbered,hidelinks]{hyperref}

\newcommand{\etal}{\textit{et al.}} 

\usepackage{color} 

\newcommand{\hkhp}[1]{\textcolor{black}{#1}}

\begin{document}
\maketitle

\begin{abstract}
Miniature-legged robots are constrained by their onboard computation and control, thus motivating the need for simple, first-principles-based geometric models that connect \emph{periodic actuation or gaits} (a universal robot control paradigm) to the induced average locomotion. In this paper, we develop a \emph{modulable two-beat gait design framework} for sprawled planar quadrupedal systems under the no-slip or pin contact constraint using tools from geometric mechanics. In this approach, we reduce standard two-beat gaits into unique subgaits in mutually exclusive, shape subspaces. These subgaits are characterized by a locomotive stance phase when limbs are in ground contact and a non-locomotive instantaneous swing phase where the limbs are reset. During the stance phase, the contacting limbs form a four-bar mechanism, and we develop the following tools to aid in designing the ensuing locomotion: (a) a vector field to generate constrained actuation, (b) the kinematics of a four-bar mechanism, and (c) stratified panels that combine the kinematics and constrained actuation to encode the net change in the system's position generated by a stance-swing subgait cycle. 
Decoupled subgaits are then designed independently using flows on the shape-change basis and are combined with appropriate phasing to produce a two-beat gait where the net displacements are added up. Further, we introduce ``scaling" and ``sliding" control inputs to continuously modulate the global trajectories of the quadrupedal system in gait time. Using this framework, we demonstrate that sprawled robot quadrupeds have a preferred direction of translation (translational anisotropy) depending on their geometry through an example trajectory. Additionally, we also demonstrate average speed, direction, and steering control using the control inputs. Thus, this framework can generate simple open-loop gait plans or gain schedules for resource-constrained robots and bring them one step closer to realizing control autonomy.
\end{abstract}

\IEEEpeerreviewmaketitle

\section{Introduction}
\label{sec:intro}

\begin{figure}[tbp]
    \centering \includegraphics[width=0.46\textwidth]{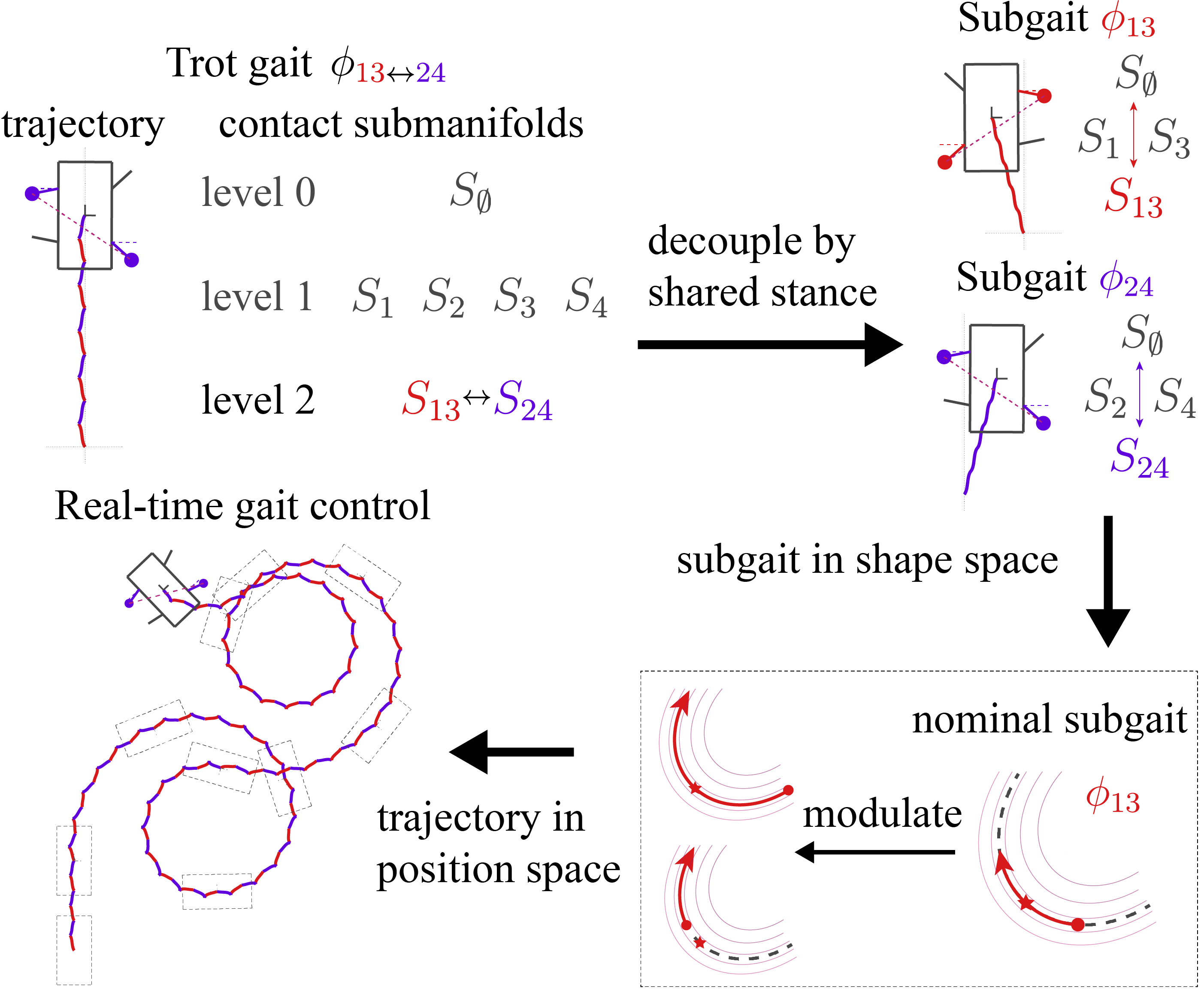}
    \caption{\small \hkhp{Overview: Two-beat quadrupedal gaits (top-left) like trot decouple into subgaits based on shared stance phase (top-right) where each subgait stance phase is a four-bar mechanism. Using geometric mechanics, we solve for the net displacement generated by integrating the \textit{stratified panel}, a measure of the effective body velocity of the system. Then, by exploiting the design symmetries, we generate steerable two-beat gaits by modulating the subgaits (bottom-right), resulting in real-time gait control (bottom-left) of rigid quadrupeds.}}
    \label{fig:overview}
\end{figure}

\IEEEPARstart{I}{nsect} scale-legged robots are becoming increasingly capable in their locomotion and have recently demonstrated rapid running \cite{goldberg_gait_2017,jayaram2020scalinghamr}, vertical and inverted climbing \cite{de2018invertedclimbHAMR} and high-speed transitions \cite{jayaram2018transition} and confined traversal \cite{jayaram2016cockroaches, kabutz2023design}.
\hkhp{In particular, these achievements have primarily relied on open-loop control using heuristic actuation schemes and experimental locomotion models \cite{goldberg2017high}, driven by the limitations of onboard power, sensing, and computation}. This approach, engineered to specific systems and scenarios, demands extensive manual tuning to achieve effective locomotion. Consequently, there is a compelling need for simple, first-principles-based models that establish connections between robot morphologies, actuation schemes, and environmental properties, which is precisely the subject of the field of geometric mechanics.

\hkhp{Geometric mechanics \cite{kelly1995geometric} has been used to model the locomotion of 
with approximate first-order dynamics using 
body deformations only
including  
principally kinematic systems \cite{zhao2022walking}, systems moving at terminal velocity \cite{hatton2013geometric} or those governed by momentum conservation laws \cite{hatton2021inertiageometry} and frictional terradynamic interactions \cite{li2013terradynamics} with the surrounding environment and to quantify average motion of periodic biological and robotic systems during swimming \cite{hatton2013geometric}, slithering \cite{hatton2010sidewinding}, crawling (by slipping) \cite{deng2023adaptive}, etc.
Legged locomotion in the context of geometric mechanics has been relatively understudied because of its hybrid dynamics or discrete changes in the underlying dynamics from discrete changes in contact. This hybrid nature when combined with the fact that multi-legged systems (four or more legs) have high-dimensional configuration spaces is an issue \cite{hatton2013gmpropel} and mandates online optimization \cite{bittner2021optimizing}, sophisticated hybrid dynamical systems theory \cite{burden2014hybrid} and numerical methods for closed-loop control \cite{kong2023hybridilqr}.}

To overcome discrete contact constraints, some researchers have relied on biological insights related to contact sequencing to generate a single \textit{hybrid local connection} \cite{chong2023optimizing} composed of local connections in each contact mode {as a function of a cyclic phase variable,}{{ a quantity that denotes the progression through a gait cycle.}} 
Then, by using the generalized Stokes' theorem, they obtain a hybrid height function that encodes displacement strength to aid in optimal gait generation. This approach has been successfully applied to analyze the locomotion of articulated robotic systems, including body bending \cite{chong2021coordbackbend}, impact of body undulation and limb contact during frictional swimming \cite{chong2023selfpropslip}, geometric mechanics in phase-based toroidal shape subspaces \cite{chong2019hierarchical}, static stability and displacement tradeoffs \cite{chong2022multileg}, and optimizing foot contact patterns \cite{chong2023optimizing}.

However, these geometric methods have not explored the properties of switching between contact states, but have focused on analyzing or optimizing standalone gaits. This has inspired us to focus on generating \emph{modulable gaits for hybrid geometric systems}. {Additionally, if the system's effort is not accounted for in hybrid gait optimization problems, it may lead to ill-conditioned gait designs as demonstrated in previous work \cite{chong2021moving}. We recently addressed this concern by studying the simplest instantiation of optimal geometric legged locomotion using a simple two-footed toy robot model with one continuous and one discrete shape \cite{prasad2023contactswitch}.}

The goal of this letter is to extend our recent work \cite{prasad2023contactswitch} to \emph{geometrically model the locomotion capabilities of sprawled, quadrupedal systems subject to no-slip contact constraints} during a two-beat gait cycle. Additionally, to enable experimental utility for our geometric gait design framework, we develop \emph{modulable geometric gaits} by exploiting the system's underlying symmetries. The stancing limbs of a quadruped during a typical two-beat gait cycle form a planar four-bar mechanism, which is analyzed in \S\ref{sec:geom_four_bar}. In \S \ref{sec:geom_quad} and \S \ref{sec:speed_dirn_steering}, we design modulable geometric trot gaits that result in average speed, course, and heading control {(or steering \cite{zhao2020multilegSlipSteer})}. Finally, in \S\ref{sec:disc}, we conclude this letter by summarizing the contributions and hinting at exciting future directions.

\section{Geometric mechanics of a four-bar mechanism}
\label{sec:geom_four_bar}
\begin{figure*}[tbp]
    \centering \includegraphics[width=0.9\textwidth]{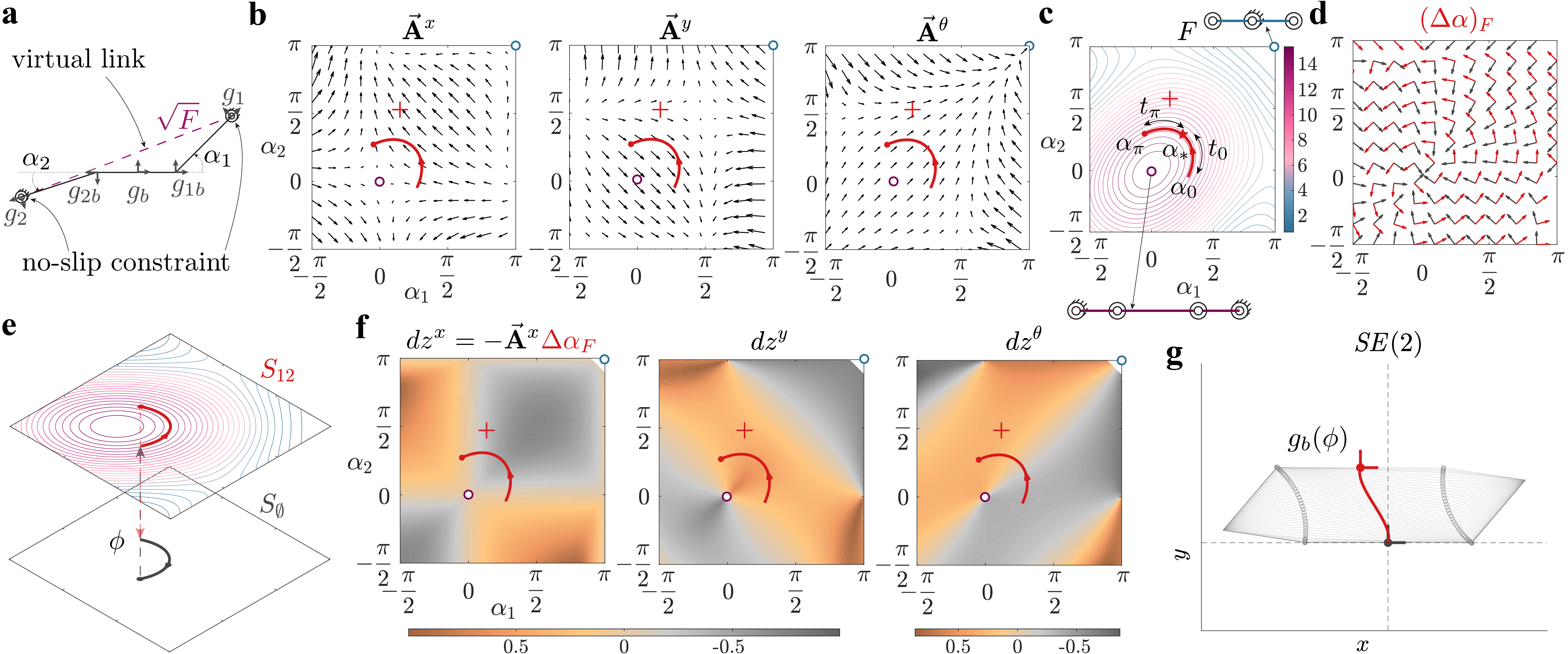}
    \caption{\hkhp{\small Geometric mechanics of a four-bar mechanism: (\textbf{a}) System with attached $\mathit{SE}(2)$ frames. (\textbf{b}) The local connection vector fields ($\vec{\mathbf{A}}$) model four-bar mechanism locomotion during the stance phase of feet $g_1$ and $g_2$. (\textbf{c}) The virtual link length (a conserved quantity under the no-slip constraint) is squared ($F$) and plotted as a function of the shape space. Its extrema (inset with configuration) are singular points in the shape space. (\textbf{d}) The derived unit shape velocity basis ($\left( \Delta \alpha \right)_F$, red quivers) corresponds to the nonslip directions tangential to the level sets of $F$. A non-slip gait example (\textbf{e}) with its stance trajectory embedded in (\textbf{b}), (\textbf{c}) (with flow parameters), and (\textbf{f}) produces the system trajectory shown in (\textbf{g}). (\textbf{f}) The stratified panels ($dz$) shown as heatmaps encode effective body velocity from an infinitesimal stance-swing gait cycle in the shape space. The embedded stance trajectory provides a visual representation of each displacement component accrued. 
    }} 
    \label{fig:4bar_gm}
\end{figure*}

As a first step towards describing the geometric mechanics of two-beat quadrupedal gaits, we consider a three-link system with distally pinned feet forming a four-bar mechanism (Fig.\ref{fig:4bar_gm}(a)). This system is the simplest instantiation of a nonslip stance phase of a typical two-beat quadrupedal gait (e.g. diagonal legs while trotting, contralateral legs while bounding, ipsilateral legs while pacing, etc. \cite{hildebrand1965symmetrical}) and represents the simultaneous stance ($S_{12}$ contact submanifold \cite{prasad2023contactswitch}). In this section, we determine the motion of this mechanism geometrically. As the first step, {we describe a typical planar four-bar linkage as a three-link system with distal ends under non-slip contact constraint, following which we obtain the geometric locomotion model and the shape (or actuation) constraint vector field related to simultaneous contact. Finally, we define a two-phase, stance-swing gait cycle using flows on the shape constraint vector field.}

\subsection{System Description}
\label{subsec:4bar__sys_des}
We describe the system's position using planar rigid-body frames (elements of $\mathit{SE}(2)$) attached at the body center, $g_b$, hip joints for each limb, $g_{1b}$ and $g_{2b}$, and foot tips, $g_{1}$ and $g_{2}$. The limb angles for each foot, $\alpha_i$, and the binary contact variables, $\beta_i$ together form the hybrid shape space of this system. The contact variable sets the current foot to a stance phase if $\beta_i = 1$ {(under non-slip contact or the pin constraint \cite{ruina1998nonholonomic} the foot is allowed to rotate once pinned, but not translate)} or a swing phase if $\beta_i = 0$. For ease of numerical analysis, we consider the system body link to be two units long and each limb to be one unit long with swing angle limits between $-\frac{\pi}{2}$ and $\pi$ radians. An illustration of this system is shown in Fig.\ref{fig:4bar_gm}(a).

\subsection{Kinematic Reconstruction using the Non-slip constraint}
\label{subsec:4bar_kin_reconstruct}
Here, we largely follow the procedure outlined in \cite{prasad2023contactswitch} to reconstruct the kinematics of a four-bar system.
We begin by determining the Jacobians that transform the system's generalized velocity (the body velocity, $\groupderiv{g}_b$ and leg-swing velocity, $\dot{\alpha}$) to the local foot velocity, $\groupderiv{g}_i$ \cite{ramasamy2019draggeometry}. For example, the {jacobian} for the first foot is given by:
\begin{align}
    J_1 = \begin{bmatrix}
            \hphantom{-}\cos \alpha_1 & \sin \alpha_1 & \sin \alpha_1 & 0 & 0\\
            -\sin \alpha_1 & \cos \alpha_1 & \cos \alpha_1+1 & 1 & 0\\
            0 & 0 & 1 & 1 & 0
        \end{bmatrix} \label{eq:J_1}
\end{align}
where the first three columns correspond to body velocity and the last two account for limb swing velocity contributions.
{{The last column is uniformly zero because swinging the second limb doesn't affect the foot velocity of the first limb.}} 
We then impose the no-slip constraint for each foot in stance to derive the kinematic reconstruction equation \cite{prasad2023contactswitch} by zeroing their translational velocity. For the multi-foot ($>2$) systems, this amounts to vertically stacking the first two rows of each stancing foot Jacobian (just the translational components isolated from the full Jacobian using a selection matrix, $C_{xy}$) to obtain a single Pfaffian constraint expression:
\beq
    \begin{bmatrix}
        C_{xy} J_1 \\
        C_{xy} J_2
    \end{bmatrix} \begin{bmatrix}
        \groupderiv{g}_b \vspace{1mm}\\
        \dot{\alpha}
        \end{bmatrix} = \left(0 \right)^{4\times1} \text{, where } C_{xy} = \begin{bmatrix}
                                1 & 0 & 0 \\
                                0 & 1 & 0
                        \end{bmatrix}    \label{eq:pfaff}
\eeq
Finally, dividing (\ref{eq:pfaff}) into two sub-matrices and then taking the pseudoinverse, as in \cite{hatton2015nonconservativity},
we obtain the local connection, $\mixedconn$
-- a linear map from the shape velocities to the body velocity, $\groupderiv{g}_b = -\mixedconn \dot{\alpha}$ defined as a function over the shape space. This linear function encodes the kinematic motion of the body frame to limb motions in a matrix form.
This matrix has two columns ($\mixedconn_1$ and $\mixedconn_2$) each corresponding to the body velocity contribution from each shape element velocity ($\dot{\alpha}_1$ and $\dot{\alpha}_2$). It has three rows of connection vector fields corresponding to each $\mathit{SE}(2)$ direction-- $(\vec{\mixedconn}^x, \vec{\mixedconn}^y, \vec{\mixedconn}^{\theta})$ (Fig.\ref{fig:4bar_gm}(b)) and are noted to be \textit{conservative} for trajectories that respect the contact constraint. When the limbs are in the swing phase $S_{\emptyset}$, we have a uniformly zero local connection matrix that represents the lack of motion during the absence of contact. 

To obtain such non-slip trajectories, we define the squared inter-foot distance, $F$, {using the $\mathit{SE}(2)$ transform} between the stancing foot frames:
\beq
    {F = {\Norm{ C_{xy} \, (g_2 \gop \inv{g_1}) }_2}^2} \label{eq:F_defn}
\eeq
Fig.\ref{fig:4bar_gm}(c) {represents the squared} inter-foot distance ($F$) as isocontours in the system's shape space. Each closed contour represents a valid domain for defining nonslip trajectories as described in the next subsection. Two distinct singular configurations occurring at extremal values of $F$ are excluded from the \emph{accessible} shape space ($\alpha_F$) because all limb motions from this state result in slipping.

\subsection{Gait definition}
\label{subsec:4bar_subgait_descrip}
A two-phase gait cycle for the four-bar mechanism has a non-slip stance phase and an unconstrained swing phase. To obtain non-slip shape trajectories, we first compute the transpose of a unit vector aligned with the gradient of F, $\transpose{\widehat{\nabla F}}$, a well-defined quantity in the tangent space of the accessible shape space ($T_{\alpha_{F}}$) and {rotate it clockwise} by \SI{90}{\degree}{ to obtain a unit vector field, $\Delta \alpha_F$, tangential to level-sets of $F$:}
\beq
    \Delta \alpha_F : \alpha_F \mapsto T_{\alpha_{F}} \suchthat \Delta \alpha_F = \begin{bmatrix}
        \hphantom{-} 0 & 1 \\
        -1 & 0
    \end{bmatrix} \transpose{\widehat{\nabla F}}(\alpha) \label{eq:allowed_path_vel}
\eeq
{This unit vector field, illustrated in Fig.\ref{fig:4bar_gm}(c) (in red), will be our basis for nonslip stance phase shape trajectories and are} defined as a \textit{flow} over $\Delta \alpha_F$ from an initial condition ($\alpha_0$):
\beq
    \alpha(t) = \Phi_{\Delta \alpha_F \, \dot{k}}^{t}(\alpha_0), \, \alpha \in \alpha_F \forall t \label{eq:noslip_flow}
\eeq
where $\dot{k}$ is a {scalar} pacing term and (\ref{eq:noslip_flow}) is the solution to the differential equation $\dot{\alpha} = \Delta \alpha_F \, \dot{k}$ at time $t$. If (\ref{eq:noslip_flow}) is used to construct a closed shape trajectory in one of the levels of $F$, this path will generate no net change in the position of the body due to the holonomic nature of the non-slip constraint.

{The locomotion-inducing stance-swing gait cycle ($\phi$) of the four-bar mechanism is defined as an instantaneous \textit{`pinning'} of the limbs, followed by} a forward flow during the stance phase ($\beta_1=\beta_2=1$, {$S_{12}$}), an instantaneous \textit{`lifting'} of the limbs, and a reversed flow during the swing ($\beta_1=\beta_2=0$, {$S_\emptyset$}) phase as shown in Fig.\ref{fig:4bar_gm}(e):
\beq
    \phi_\tau = \begin{bmatrix}
        \alpha \\
        \beta
    \end{bmatrix}_\tau = \begin{cases}
        \begin{bmatrix}
            \Phi_{\Delta \alpha_F \, \dot{k}}^{\tau}(\alpha_0) \\
            {\transpose{[1, 1]}}
        \end{bmatrix}, \:\: \tau \in [0, \pi] \\
        \begin{bmatrix}
            \Phi_{\Delta \alpha_F \, \dot{k}}^{\pi - \tau} (\alpha_{\pi})  \\
            {\transpose{[0, 0]}}
        \end{bmatrix}, \:\: \tau \in [\pi, 2\pi]
    \end{cases} \label{eq:subgait_flow}
\eeq
where $\tau \in \sphere^1$ is the gait phase (or kinematic phase \cite{revzen2013instantaneous}).

The initial and final conditions of the stance phase (and vice versa for the swing phase) from (\ref{eq:subgait_flow}) are $\alpha_{\pi} = \Phi_{\Delta \alpha_F}^{\pi}(\alpha_0)$ and $\Phi_{-\Delta \alpha_F}^{\pi}(\alpha_{\pi})$. 
For a forward path of length $t$, a constant unit pacing is $\dot{k} = \frac{t}{\pi}$. 
The above definition explicitly defines a reversed flow during the swing phase to conserve {the $F$ level-set} in each contact submanifold $S_{12}$ and $S_{\emptyset}$ (Fig.\ref{fig:4bar_gm}(e)). Other swing paths (say a straight-line return) can be defined as perturbations on the swing definition in (\ref{eq:subgait_flow}) {and} is outside the scope of this work.

Finally, the body trajectory ($g_b$) during the gait cycle is obtained by {flowing along} the body velocity ($\groupderiv{g}_b$) in the rest frame ($\leftliftedaction{e}{g_b} \, \groupderiv{g}_b$, but shown as $g \, \groupderiv{g}_b$ for brevity):
\begin{align}
    \dot{g}_b(\tau) &= -g \, \groupderiv{g}_b = -g \mixedconn \, \Delta \alpha_F \dot{k}, \:\: \tau < \pi \label{eq:pulledback_body_velocity} \\
    g_b(\tau) &= \begin{cases}
        \Phi_{\dot{g}_b(\tau)}^{\tau}( \, g_b(0) \, ), \:\: \tau \leq \pi \\
        g_b(\pi), \:\: \pi < \tau < 2\pi
    \end{cases} \label{eq:body_trajectory}
\end{align}

\subsection{Stratified Panel}
\label{subsec:4bar_strat_panel}
We determine the average motion accrued by a planar four-bar mechanism by obtaining the stratified panel over a stance-swing gait cycle. 
The stratified panels (or discrete curvature functions) \cite{prasad2023contactswitch} are a measure of a hybrid system's {effective body velocity in a contact-switching gait cycle between any two contact states. 
In the case of a four-bar system, the contact switches occur between the stance and swing phases in the contact submanifolds $S_{12}$ and $S_{\emptyset}$.} These are analogous to constraint curvature functions \cite{hatton2015nonconservativity,ramasamy2019draggeometry, chong2021coordbackbend,chong2021moving} for systems {under a single locomotion constraint}.
To compute this for a hybrid system, we consider an infinitesimal stance-swing gait cycle ($\hat{\phi}$) with unit pacing centered around a limb position ($\alpha$).

From (\ref{eq:subgait_flow}), 
the {corresponding} infinitesimal net displacement ($z_{\hat{\phi}}$) {is given by:
}
\beq
z_{\hat{\phi}} = - \mixedconn \Delta \alpha_F (\alpha) \, d\tau = {dz_{\hat{\phi}}} \, d\tau, \label{eq:strat_panel_4bar}
\eeq
{In (\ref{eq:strat_panel_4bar}), the} \emph{stratified panel} ($dz$) {is represented as a function of the shape space ($\alpha$) and defined as} a product between the local connection ($\mixedconn$) and the path constraint vector field ($\Delta \alpha_F$). 
For four-bar systems, these (Fig.\ref{fig:4bar_gm}(g)) show interesting symmetries about certain shape space axes. If we restrict our available limb positions ($\alpha \in \pm\frac{\pi}{2}$), the translational panels ($x$ and $y$-panels) exhibit symmetry about the $\alpha_1 = \alpha_2$ and $\alpha_1 = -\alpha_2$ axes, whereas the rotational panel (or $\theta$-panel) exhibits skew-symmetry about the $\alpha_1 = \alpha_2$ axis. The properties of position space and these panels taken together play an important role in gait construction (\S\ref{subsec:4bar_trans_gait_construction}).

An example stance trajectory is shown in Fig.\ref{fig:4bar_gm}(f). Its anticlockwise orientation (denoted with a plus sign) implies that the shape velocity is aligned with $\Delta \alpha_F$ and the illustrated panel heatmap (Fig.\ref{fig:4bar_gm}(g)). If the orientation is reversed, then the panel estimates have to be negated accordingly. The net displacement ($z_{\phi}$) produced by this gait cycle is obtained from (\ref{eq:body_trajectory}) by setting $\tau = 2\pi$. 
Particularly, we note that this gait cycle does not generate a net rotation relative to the body frame and is therefore useful for generating pure translation gaits (\S \ref{subsec:4bar_trans_gait_construction}).

\subsection{Gait construction}
\label{subsec:4bar_trans_gait_construction}
The system's $\mathit{SE}(2)$ position space exhibits noncommutativity between its rotational ($\mathit{SO}(2)$) and translational ($\euclid^2$) subspaces \cite{hatton2015nonconservativity}. However, the rotational subspace is commutative implying that paths in the $\theta$-panel can be directly integrated to obtain the net rotation induced by gaits. 
We exploit this feature along with the skew-symmetry exhibited by the $\theta$-panel to generate pure \emph{translation gaits} that accrue zero net rotation.

For this, we choose a reference point $\alpha_{*}$ along the \hkhp{zero contour of the $\theta$-panel that happens to be the $\alpha_1 = \alpha_2$ line for a desired $F$-level to operate in.} Then, given a desired path length $t$, the initial ($\alpha_0$) and final states ($\alpha_{\pi}$) of the stance path can be obtained using through nonslip flows from $\alpha_{*}$:
    \begin{align}
        \alpha_0 = \Phi_{\Delta \alpha_F}^{\pm\frac{t}{2}}(\alpha_{*}); 
        \alpha_{\pi} = \Phi_{\Delta \alpha_F}^{\mp\frac{t}{2}}(\alpha_{*}) \label{eq:ptf2fc},
    \end{align}
\hkhp{Because $\alpha_{*}$ is on the zero-contour of the $\theta$-panel for an ipsi-laterally symmetric system, 
the net rotation generated by the stance-path segments of length $\frac{t}{2}$ from $\alpha_0$ to $\alpha_{*}$ and from $\alpha_{*}$ to $\alpha_{\pi}$ cancel out each other.}
More generally, the initial and final states from the $\alpha_{*}$ are defined with path lengths $t_0$ and $t_{\pi}$ with the total stance path being $t = t_0 + t_{\pi}$. For arbitrary values, the rotational skew-symmetry is broken and results in gaits that also generate rotation. This idea will be explored to enable steering for quadrupedal robots in \S \ref{subsec:control_knob}.

\section{Geometric mechanics of planar quadrupeds}
\label{sec:geom_quad}
\begin{figure}[tbp]
    \centering \includegraphics[width=0.45\textwidth]{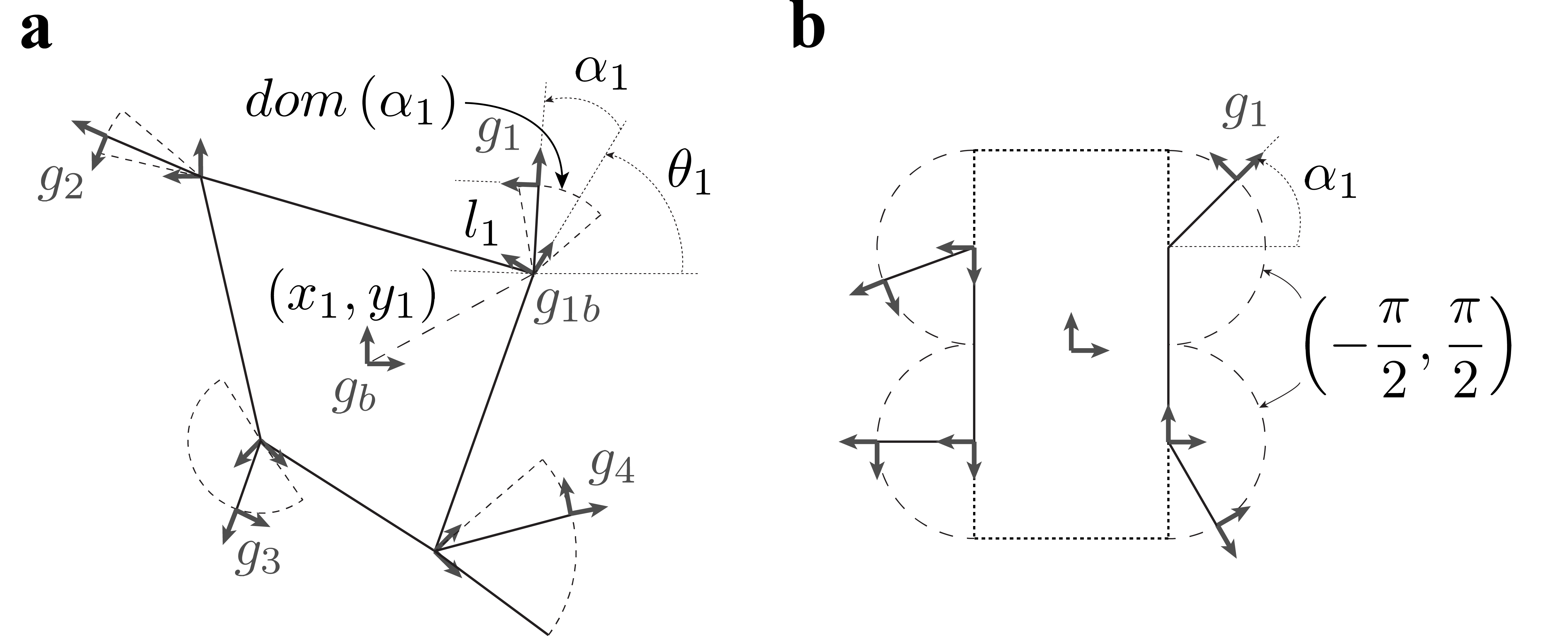}
    \caption{\small (\textbf{a}) A generic planar quadrupedal system is defined using parameters for each leg module. For our choice, we obtain a (\textbf{b}) fiducial rigid quadrupedal system used in \S\ref{sec:geom_quad}.}
    \label{fig:quad_defn}
\end{figure}
In this section, we generate two-beat gaits for rigid quadrupedal systems by combining the motion of the body in two stance phases where the geometry of each stance phase representing a \textit{subgait} is dictated by a four-bar mechanism as described above. 
We start by detailing the rigid quadrupedal system and formally define a two-beat trot gait cycle as a function of the two subgaits formed in the shape subspace of limbs with a shared stance phase. We then reframe the subgait flow definition as a two-parameter family to realize the ability to continuously modulate geometric gaits as desired and demonstrate this capability by presenting three control strategies to vary speed, direction, and steering.

\subsection{System description}
\label{subsec:quad_sys_des}
A generalized planar sprawled quadrupedal system (Fig.\ref{fig:quad_defn}(a)) is defined by a hip frame ($g_{ib}$ frame) for each leg, leg length ($l_i$), leg swing angle ($\alpha_i$), foot location ($g_{i}$), and contact state ($\beta_i$) respectively. For simplicity and intuitive description of our methodology, we choose geometric parameters (body-leg ratio, leg locations relative to the center of mass, etc.) for our {fiducial} system to emulate {rigid sprawled quadrupeds like} Harvard Ambulatory MicroRobot (HAMR), a platform co-developed and used by our group extensively \cite{goldberg_gait_2017,doshi2019effective,jayaram2020scalinghamr}.
The body of {our fiducial} sprawled system (Fig.\ref{fig:quad_defn}(b)) is symmetric about both body axes with a \SI{90}{\degree} bidirectional swing range {and modeled to have a total body length of four units. Each hip frame has unit translational offsets in the $x$ and $y$ directions, and zero rotational offsets from the central body frame.}

\subsection{Two-beat gait construction}
\label{subsec:quad_trot}
For generating two-beat gaits in rigid quadrupeds, we leverage the insight from biological locomotion \cite{hildebrand1965symmetrical} that typical two-beat gait cycles can be uniquely divided into subgaits based on shared stance phases. As an example, we choose to demonstrate this by {considering} a trot gait where diagonal legs alternate relative to each other. For a trot gait ($\phi_{13 \leftrightarrow 24}$), the first subgait ($\hat{\phi}_{13}$) {is defined in the reduced shape subspace, $B_{13}$ during the stance contact submanifold $S_{13}$,} and the second subgait ($\hat{\phi}_{24}$) {is defined in $B_{24}$ during $S_{24}$. The reduced shape subspace $B_{ij}$ extracts the shape variables corresponding to limbs `$i$' and `$j$' and helps us directly use the gait definition from (\ref{eq:subgait_flow}).}
This approach can be adapted to design other two-beat gaits such as bound ($\phi_{12 \leftrightarrow 34}$) and pace ($\phi_{23 \leftrightarrow 41}$) with the contralateral and ipsilateral legs in shared stance.

\hkhp{Moving forward, we shall only consider trot gaits with \SI{50}{\%} duty cycle {and} no overlap between {the stance phases of} subgaits to limit the current manuscript scope}. This provides the following trot gait description (by offsetting the phase between subgaits by half a cycle, $\pi$) and net displacement:
\begin{align}
    \phi_{13 \leftrightarrow 24}(\tau) = \hat{\phi}_{13}(\tau) + \hat{\phi}_{24}(\pi + \tau) \nonumber\\
    z(\phi_{13 \leftrightarrow 24}) = z(\hat{\phi}_{13}) + z(\hat{\phi}_{24}) \label{eq:trot_inGC}
\end{align}
Due to the holonomic nature of each stance phase, the net displacement of the system over a gait cycle is the sum of the displacements from each subgait as long as there is no overlap between the stance phases. \hkhp{In future work, we can relax the non-slip assumption to provide more insight into the system's motion during such overlap.}

\begin{figure}[tbp]
    \centering \includegraphics[width=0.45\textwidth]{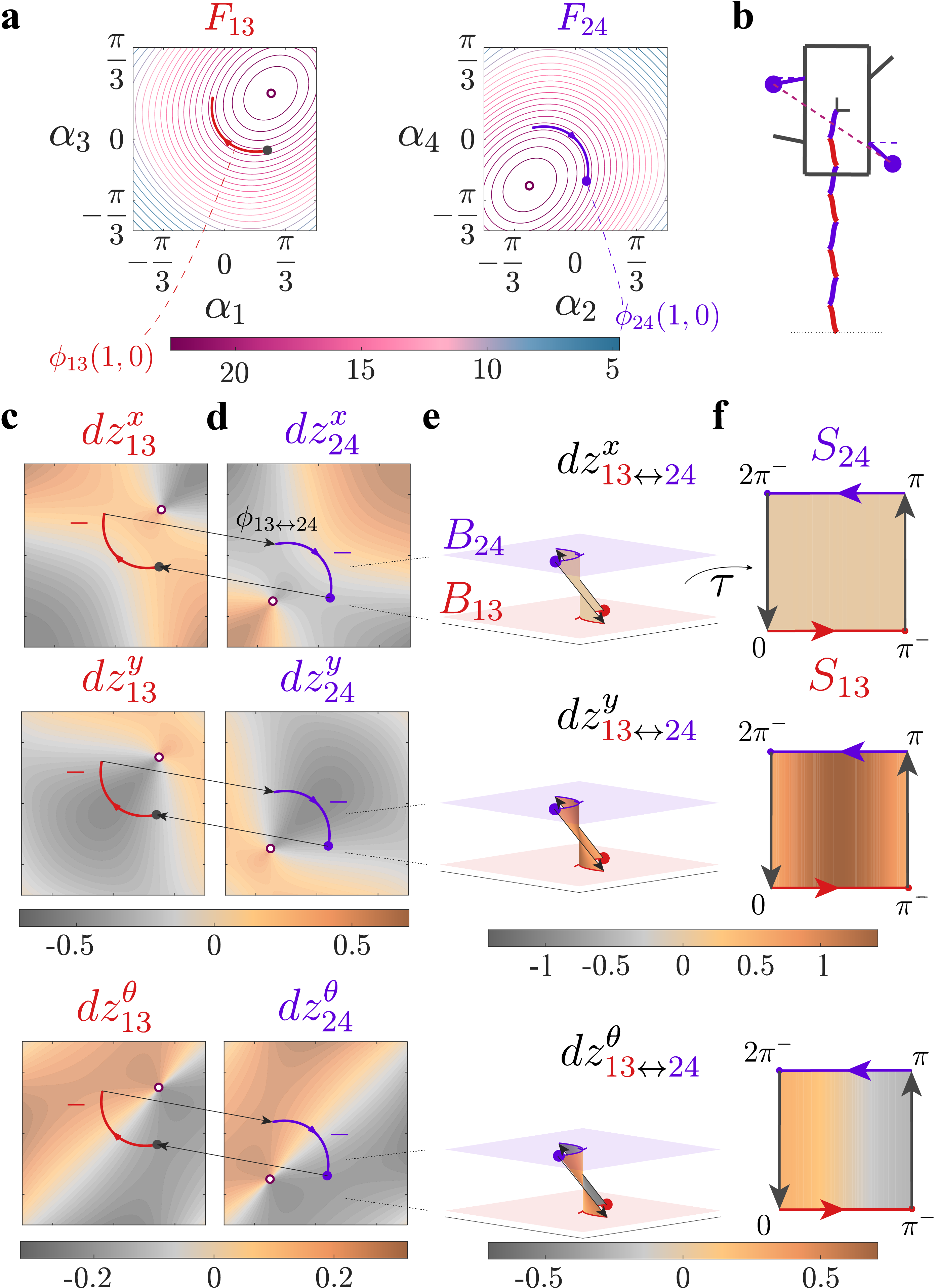}
    \caption{\small \hkhp{Geometric trotting: Forward displacing gait constituted by subgaits ($\phi_{13}$ and $\phi_{24}$) with highlighted stance phases with negative orientation. (\textbf{a}) The contours of $F_{ij}$ in each shape subspace represent the domain for nonslip shape paths. (\textbf{b}) {The corresponding robot trajectory. 
    } The stratified panels, $dz$ of the two contact states (\textbf{c}) and (\textbf{d}), (refer to (\textbf{a}) for axes information) correspond to anticlockwise trajectories around the inset singularity. The full stratified panel between the two stance paths in reduced shape spaces (\textbf{e}) and as a function of gait phase, $\tau$ in (\textbf{f}).}}
    \label{fig:trans_trot}
\end{figure}

\hkhp{Using gait definition (\ref{eq:trot_inGC}) and translational gait construction methodology (\S \ref{subsec:4bar_trans_gait_construction}), we construct an example forward-displacing trot gait (Fig.\ref{fig:trans_trot}) with the reference point being the origin of each reduced shape space, $B_{ij}$ $(\alpha_i, \alpha_j) = (0, 0)$.
Due to the symmetric nature of the system and by extension, the shape subspace, we choose negative flow path lengths $(t_0, t_{\pi}) = (-0.8, -0.8)$ for both subgaits to obtain clockwise stance paths that generate equal $y$-displacements, and net-zero $x$ and $\theta$-displacements.} 

This approach to estimating gait motion through a piecewise combination of subgaits based on shared stance is distinct from our previous methodology of gait construction using stratified panels \cite{prasad2023contactswitch} for hybrid systems, but can be easily reformulated as below.
The effective two-beat body velocity or the \emph{two-beat stratified panel} is defined through a temporal sequence of events: a first stance phase ($S_{ij}$), an infinitesimal forward contact switch ($S_{ij} \rightarrow S_{kl}$), a second stance phase ($S_{kl}$), and an infinitesimal backward contact switch ($S_{ij} \leftarrow S_{kl}$). For our symmetric forward-displacing trot gait, these overall panels are computed at gait phase $\tau$ as the sum of the subpanels at phases $\tau$ and $-\tau$:
\beq
    dz_{13 \leftrightarrow 24} = dz_{13}(\tau) + dz_{24}(-\tau) \label{eq:trot_2bpanel}
\eeq
These panels (Fig.\ref{fig:trans_trot}(e)) form warped surfaces between the stance trajectories $\hat{\phi}_{13}$ and $\hat{\phi}_{24}$ in $B_{13}$ and $B_{24}$. However, in gait phase space, $\tau$, these complex surfaces can be unwrapped and redrawn as rectangular stratified panels  (Fig.\ref{fig:trans_trot}(f)) where each vertical slice encodes the effective velocity of an infinitesimal switch between the stance phases similar to our earlier work \cite{prasad2023contactswitch}. Although it provides similar insights in the net displacement accrued, the two-beat stratified panels (Fig.\ref{fig:trans_trot}(e) and (f)) have to be constructed separately for every combination of subgaits and therefore is not practical for gait generation for multilegged (or high dimensional hybrid) systems. Moving forward, we shall stick to using the subpanels for steerable gait generation.

{\subsection{Flow-control of subgaits}
\label{subsec:control_knob}}
\hkhp{As outlined previously in \S\ref{subsec:4bar_trans_gait_construction} and \S\ref{subsec:quad_trot}, it is useful to take advantage of the integrability of the rotational subspace to generate pure translational gaits using stratified panels. These translational gaits apart from being easy to construct, are also useful as a fiducial starting point to generate more complex gaits. Because the stratified panels are continous and well-behaved around the stance path, continous control of subgaits can be achieved as follows:
\begin{enumerate}
    \item Firstly, in translational gaits, the displacements generated by each subgait is a vector 
    in the 
    locomotion plane. Scaling these displacement vectors can provide average course or speed control of the system.
    This inspires \emph{scaling flow input} that amplifies the path length of a subgait by applying equal scaling to the forward and reverse flow times $t_0$ and $t_{\pi}$ while preserving the reference point $\alpha_{*}$. Incorporating this input ($u_1 \in \euclid$) into (\ref{eq:ptf2fc}), we obtain:
        \begin{align}
            \alpha_0 = \Phi_{\Delta \alpha_F}^{u_1 t_0}(\alpha_{*}); \, 
            \alpha_{\pi} = \Phi_{\Delta \alpha_F}^{-u_1 t_{\pi}}(\alpha_{*}) \label{eq:ptf2fc_arb_wU1}
        \end{align}
    \item Secondly, where turning is viable, we introduce a \emph{sliding flow input} that provides a unidirectional affine shift to the flow times $t_0$ and $t_{\pi}$. Incorporating this input ($u_2 \in \euclid$) into (\ref{eq:ptf2fc_arb_wU1}), we obtain:
        \begin{align}
            \alpha_0 = \Phi_{\Delta \alpha_F}^{u_1 t_0 + u_2}(\alpha_{*}); \, 
            \alpha_{\pi} = \Phi_{\Delta \alpha_F}^{-u_1 t_{\pi} + u_2}(\alpha_{*}) \label{eq:ptf2fc_arb_wU12}
        \end{align}
\end{enumerate}}
Incorporating (\ref{eq:ptf2fc_arb_wU1}) and (\ref{eq:ptf2fc_arb_wU12}) in (\ref{eq:subgait_flow}) provides two, two-parameter family of subgaits, $\hat{\phi}_{13}(\vec{u}^{13})$ and $\hat{\phi}_{24}(\vec{u}^{24})$, which when combined with (\ref{eq:trot_inGC}) provides a four-parameter family of modulable geometric trot gaits, $\phi_{13 \leftrightarrow 24}(\vec{u}^{13}, \vec{u}^{24})$.

\begin{figure}[tbp]
    \centering \includegraphics[width=0.45\textwidth]{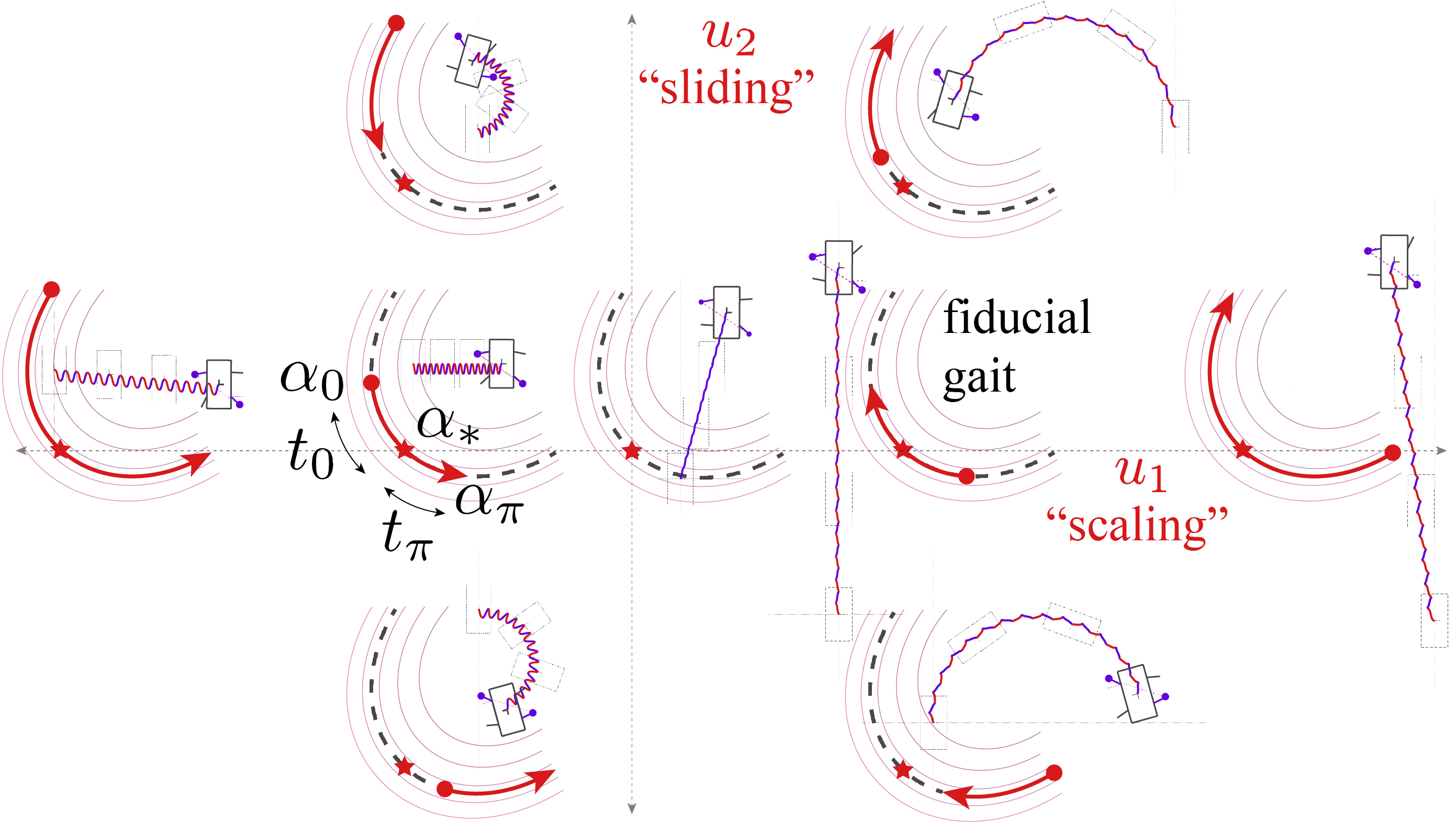}
    \caption{\small {Flow control of $\hat{\phi}_{13}$ while keeping $\hat{\phi}_{24}$ fixed: All the gait cycles are derived by modulating the flow control inputs of the fiducial forward gait cycle. The scaling input modifies the path length whereas the sliding input modifies the path location relative to the reference point ($\alpha_{*}$). 
    }}
    \label{fig:control_knobs}
\end{figure}

To illustrate the effect of these two flow inputs on locomotion (Fig.\ref{fig:control_knobs}), we consider the forward-displacing trot gait from \S \ref{subsec:quad_trot} with unit scaling ($u^{13}_1 = u^{24}_1 = 1$) and zero sliding ($u^{13}_2 = u^{24}_2 = 0$) inputs as our fiducial trajectory. Further, we choose to apply varying flow inputs (one at a time) to the subgait cycle, $\hat{\phi}_{13}(\transpose{[u_1^{13}, u_2^{13}]})$, while keeping the other subgait cycle, $\hat{\phi}_{24}$ constant.

\begin{figure*}[tbp]
    \centering \includegraphics[width=0.9\textwidth]{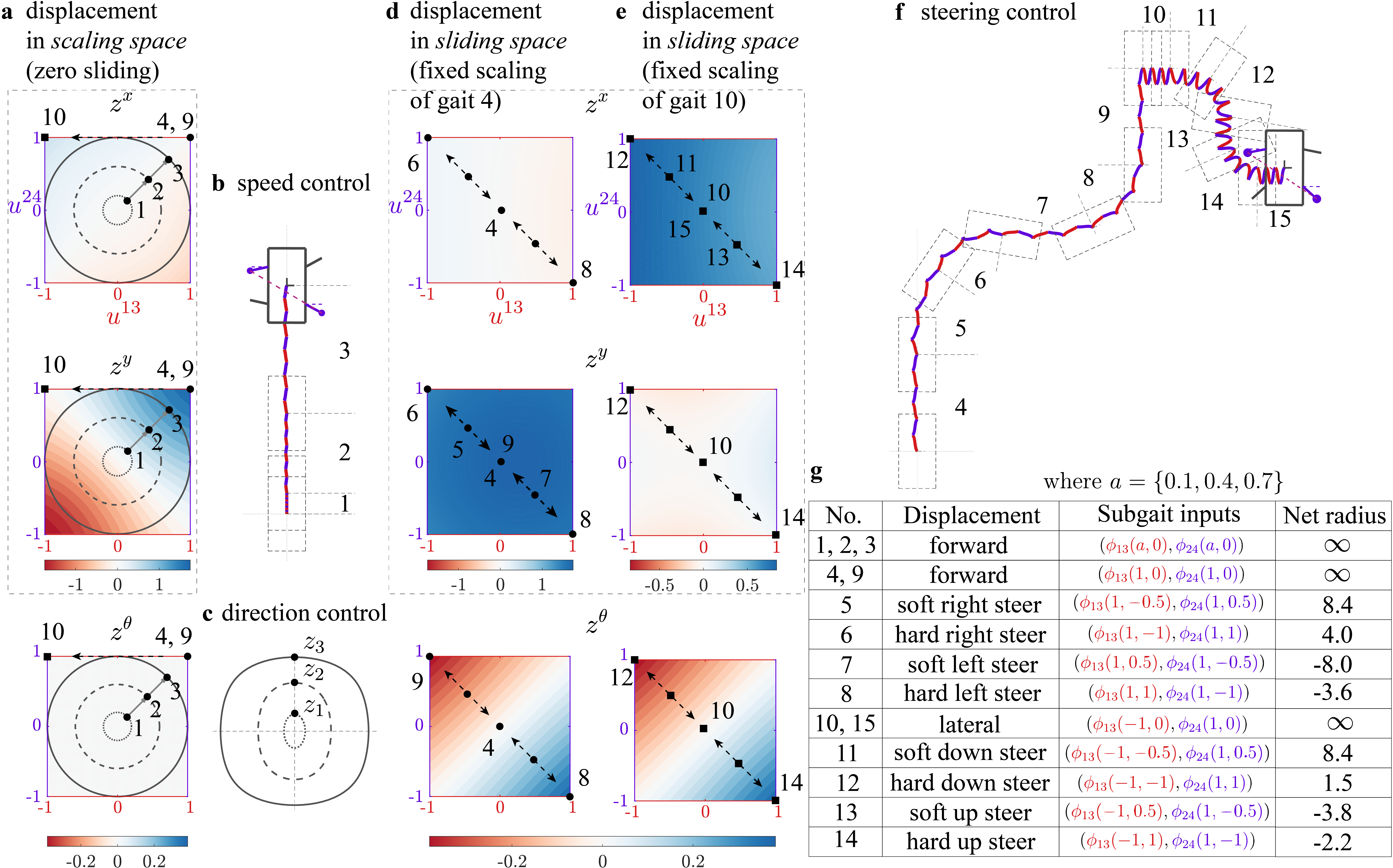}
    \caption{\small {\hkhp{Modulable} Geometric Trotting: (\textbf{a}) The net displacement generated by a two-beat trot gait cycle as a function of the scaling input subspace (with zero sliding). (\textbf{b}) Speed control along the heading direction is achieved by uniformly scaling gains along the $u^{13}_1 = u^{24}_1$ axis. (\textbf{c}) Varying the scaling parameters along a circular path provides continuous control over the displacement direction. Steering control: For the gaits $4$ and $10$, the scaling inputs are fixed and the net displacement is plotted as a function of the sliding parameter subspace in (\textbf{d}) and (\textbf{e}). These displacement fields depict the Lie-bracket effect provided by the sliding gains. Gaits $4$--$9$ provide steering about a forward trot, whereas $10$--$15$ provide lateral steering. List (\textbf{g}) provides additional information on the gait cycles.}}
    \label{fig:path_control}
\end{figure*}

\section{Average locomotion control}
\label{sec:speed_dirn_steering}
Modulating subgait $\hat{\phi}_{13}$ in \S \ref{subsec:control_knob} provided illustrative examples of the potential for continuous (on a per cycle basis, and therefore, \textit{average}) changes to heading ($\theta$), forward ($x$) and lateral ($y$) displacements during quadrupedal trotting. 
{In this section, we expand on these insights to control three different gait-cycle averaged locomotion scenarios generated by a trot gait: course of the system, speed along a course, and current system heading. This is visualized by considering the net displacement per trot cycle as a function of the scaling ($u_1^{13}, u_1^{24}$) or the sliding subspaces ($u_2^{13}, u_2^{24}$) respectively.}
For ease of illustration and interpretation, we choose to modulate these inputs independently and note that more complex control may be achieved by varying them simultaneously. 
We designate the origins {of the reduced shape space as reference points for each subgait} (origin of $B_{ij}$ for $\hat{\phi}_{ij}$) with flow times: $(t_0, t_{\pi}) = (0.8, 0.8)$ and limit both scaling and sliding inputs to the range $[-1, 1] \in \mathbb{R}$. 

To realize average speed control along a desired course, we determine {the net system displacements as a function of the scaling subspace (Fig.\ref{fig:path_control}(a)) for the chosen forward-displacing trot gait (\S \ref{subsec:quad_trot}, Fig.\ref{fig:trans_trot}) while prescribing the sliding inputs to be zero. As before, the skew symmetry of the $\theta$-panels ensures that the heading remains unaffected by the scaling inputs. The $x$-displacement field increases with increasing $u_1^{24}$ and decreasing $u_1^{13}$ along $u_1^{13} = -u_1^{24}$ axis with the zero contour along $u_1^{13} = u_1^{24}$ axis. Similarly, the $y$-displacement field increases with increasing scaling along $u_1^{13} = u_1^{24}$ axis with the null contour along $u_1^{13} = -u_1^{24}$. Due to the relative orthogonality between the null contours, we can continuously modulate course (forward or backward, right or left) without changing the heading using scaling inputs along either the $u_1^{13} = u_1^{24}$ axis or $u_1^{13} = -u_1^{24}$ axis respectively. The gains along these axes map to a continuous range of bidirectional displacements thereby providing \emph{average speed controlled trot gaits along a desired course}. For example, Fig.\ref{fig:path_control}(b) depicts forward trajectory with increasing average speed every five gait cycles generated by labeled gaits $1$ (slowest) through $3$ (fastest).} 
However, due to the geometry of the system where lateral $x$-sprawled limbs of the rigid quadrupedal system provide mobility primarily along the longitudinal $y$-direction, we observe \emph{translational anisotropy} between the magnitudes of the $x$ and $y$ displacement fields (Fig.\ref{fig:path_control}(a)). 
To highlight this displacement disparity, we consider three circular sets of scaling inputs defined as follows:
\beq
    (u_1^{13}, u_1^{24}) = a(\cos{\delta}, \sin{\delta}), \,\, \delta \in \sphere^{1} \label{eq:dirn_gains}
\eeq
where $a \in \{0.1, 0.4, 0.7\}$. Each set contains the labeled gaits $1$ to $3$ (Fig.\ref{fig:path_control}(a)) and the corresponding closed contours in the position space display the course of the system (Fig.\ref{fig:path_control}(c)). Extracting gains for a specific course at different circular level sets provides a general framework for average speed control along a specific course without changing the heading.

To realize average heading control, we determine the net system displacements as a function of the sliding subspace ($(u_2^{13}, u_2^{24})$) for two chosen trot gaits - a forward-displacing (Fig.\ref{fig:path_control}(d)) and a right-displacing (Fig.\ref{fig:path_control}(e)) defined by fixed scaling inputs $(1, 1)$ and $(-1, 1)$ respectively.
The sliding subspace largely preserves the translational displacements of both gaits, while the $\theta$-displacement field provides a continuous bidirectional range of net rotations along the $u_2^{13} = -u_2^{24}$ axis. To demonstrate steering, we provide a composite robot trajectory where the first half demonstrates longitudinal steering through labeled gaits $4$ through $9$, and the second half demonstrates lateral steering with labeled gaits $10$ through $15$. Furthermore, to provide path-curvature-based motion planning utility {to our steering method}, we compute the net turning radius ($r_{\phi}$) as a function of the net displacement:
\beq
    r_{\phi} = \frac{\sqrt{(z^x)^2 + (z^y)^2}}{2 \sin{\frac{z^{\theta}}{2}}} \label{eq:turning_radius}.
\eeq

\hkhp{While we have studied modulability of trot gaits, doing so optimally requires trading off the system's effort to the incurred displacement as described in Deng \etal \cite{deng2022maneuverability}. 
By combining current results with formulations in our prior work \cite{prasad2023contactswitch, deng2022maneuverability, ramasamy2016soap}, we can perform a similar analysis with an actuation-relevant metric in the shape space to derive optimal maneuverability in a neighborhood around optimal gaits.
}

\section{Discussion}
\label{sec:disc}
In this work, we provide a geometric modeling based two-beat gait design framework for planar rigid-bodied quadrupedal robots subject to the no-slip constraint. 
This subgait-driven description of a composite gait cycle or a \emph{gait-coordinate system} is directly built on previous work on simpler hybrid systems \cite{prasad2023contactswitch} with one continuous shape and one discrete shape. 
Because the subgaits are decoupled with no stance overlap, they can be designed in isolation and their net displacements add up. With trotting as an example, we designed translational gaits (\S \ref{subsec:quad_trot}) by canceling the rotation generated by subgaits. Therefore, to generate rotation for steering 
in a principled manner, we introduced the scaling and sliding inputs or \textit{'control knobs'} to continuously modulate the subgaits. Finally, taking advantage of the \hkhp{full gait control space}, we demonstrated average speed control, direction control, and steering control (\S \ref{sec:speed_dirn_steering}) and showed that laterally sprawled limbs primarily provide longitudinal mobility for typical quadrupeds.

\hkhp{In contrast to existing work on geometric legged locomotion
where slipping-contact is modeled
through Resistive Force Theory (RFT) \cite{li2013terradynamics}
and solved through numerical optimization, 
our approach
provides a simpler framework for nonslip contact directly as a pseudo-inversion-ready Pfaffian constraint for multilegged stancing systems through  group operations. Our approach also provides exhaustive modulable gaits or gain-schedules that can be computed offline and cached for online use in geometric gait planners for high maneuverability in the horizontal plane.
An exciting extension would be to obtain optimal geometric gait planners by combining our gait parameterizations with optimal maneuverability methods \cite{deng2022maneuverability}.}

\hkhp{For arbitrary systems, scaling and sliding inputs differ only quantitatively from the cases considered here, but may have restrictions on the range of accessible average locomotion or be hard to identify. 
However, for a system with longer hind limbs, the scaling input would relatively scale the forward and backward path-lengths ($t_{0}$, $t_{\pi}$) based on the front-hind limb-length ratio to ensure pure translation without rotation. Similarly, the sliding input should preserve the bidirectional steering capabilities without compromising the average course velocity. Therefore, to aid in the inverse problem of designing a system with sufficient maneuverability in the locomotion plane, a comprehensive \emph{mobility} framework is needed to connect system parameters to stratified panels.}

\hkhp{That said,} our framework is suited for 
control of 
resource-constrained legged robots \cite{jayaram2020scalinghamr, kabutz2023mclari}.
that currently rely on 
phase control \cite{doshi_phase_2017}, heuristic tuning of biological gaits \cite{goldberg2017high}, contact-ignored LQR control \cite{doshi2019effective}, and contact-implicit trajectory optimization framework \cite{doshi2019contact}.
In scenarios such as climbing \cite{de2018invertedclimbHAMR}, our methods could provide an exact solution \hkhp{because} slipping (or violation of the no-slip constraint) can be fatal for the robot. 
\hkhp{Similarly, using our framework, larger state-of-the-art robotic systems \cite{hutter2016anymal} can benefit from more efficient geometric gait planners to save valuable resources during simple navigation tasks and obtain simple gait primitives or \emph{seed gaits} for online trajectory optimization techniques \cite{kong2023hybridilqr, doshi2019contact} during complex terrain locomotion.}

As immediate follow-up steps, we will further investigate the efficacy of geometric models based on non-slip contact and their predictive capabilities. Next, we will test our open-loop gait plans on robotic platforms at different scales to experimentally validate the scale-agnostic nature of our results and extend this work to model robots \cite{mcclintock2021fabrication, kabutz2023design, kabutz2023mclari, hedrick2024femtosecond} with deformable body shapes. Finally, leveraging recent advances in efficient model-predictive controllers (MPC) targeting resource-constrained systems \cite{alavilli2023tinympc}, our geometric framework \hkhp{can be adapted} to achieve control autonomy for tiny-legged microrobots to enable animal-like performances \cite{burden2024animals}.

\bibliographystyle{ieeetran}
{\small
\bibliography{IEEEabrv, my_references}}
\end{document}